# Gland Segmentation in Histopathology Images Using Random Forest Guided Boundary Construction

Supervisor                 Angshuman Paul

Rohit A.P.                  IISc Bangalore
Salman Siddique Khan        NIT Rourkela
Kumar Anubhav               Jadavpur University

# Gland Segmentation in Histopathology Images Using Random Forest Guided Boundary Construction

Rohith AP[1], Salman S. Khan[2], Kumar Anubhav[3]

*Abstract*—Grading of cancer is important to know the extent of its spread. Prior to grading, segmentation of glandular structures is important. Manual segmentation is a time consuming process and is subject to observer bias. Hence, an automated process is required to segment the gland structures. These glands show a large variation in shape size and texture. This makes the task challenging as the glands cannot be segmented using mere morphological operations and conventional segmentation mechanisms. In this project we propose a method which detects the boundary epithelial cells of glands and then a novel approach is used to construct the complete gland boundary. The region enclosed within the boundary can then be obtained to get the segmented gland regions.

*Index Terms*—Random Forest, Gland Segmentation, Histopathology Image Analysis.

## I. INTRODUCTION

Colon cancer is one of the leading causes of death in the world. Proper treatment of cancer requires a grading to be done on the stage of cancer. An important method of grading is to know the morphology of the glands present in a tissue histopathology image. Thus gland segmentation is the first step of cancer grading. Through gland segmentation, we mark out the glands from the images. The images which are used here have been stained using Hematoxylin and Eosin. The gland regions are distinctly visible and consists of nuclei, cytoplasm and lumen. Hence a gland segmentation algorithm should capture these components in the results. Fig 1 shows a sample gland tissue histology image with various components labelled. The epithelial cells are darker and are shown by 'E' label. Lumen is the central white region marked by 'L'. The goblet cells surround the central white lumen and is marked by 'G'. The entire gland is embedded in stroma labelled by 'SN'. There are also some other epithelial cells present in the stroma.

To detect colon cancer from a digitized tissue slide, the pathologist uses (a) structural information – like glands in a cancer region have structural properties such as nuclei abundance and lumen size different from glands in a normal region and contextual information – cancer glands typically cluster into groups and are of similar shape and size, whereas shape and size of normal glands vary widely.

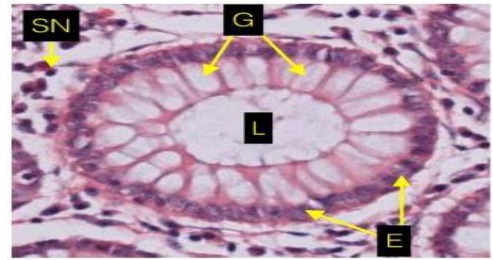

Fig 1: A colon histology image showing various components (epithelial cell or E, stromal nucleus or SN, lumen or L, goblet cell or G)

There is significant increase in the volume of colon screening tests. The shortage of pathologists in the face of the increased volume is a strong driver to develop an automated procedure for classification of gland as benign or malignant. Thus the first step in this process would be to get the segmented glands properly.

The glands which are present in tissue images do not have any uniform shape, size, color or texture. The variation in the structure of the glands is very large as we progress from benign cases to the highly malignant ones. This makes the segmentation process very tricky as one has to take into account various possible characteristics and features of glands. A method which would work well for benign glands may produce absurd results for the malignant ones. Hence an acceptable solution would be one which gives reasonable results for all the cases.

A feature which is common to all the glands is that the glands are surrounded by a layer of epithelial cells. Thus if we could successfully extract the boundary epithelial cells of a gland, it would be possible to know the shape of the gland and segmentation could be made possible from this information. By making use of this observation we propose a novel method for segmenting the glands. First, we find out all the epithelial cells in the tissue image using simple thresholding by Otsu's method[1]. Next we classify the epithelial cells as either belonging to the gland boundary or as a stromal epithelial cell using Random Forest. In the third step we connect together all the boundary epithelial cells which belong to one gland so that we get the closed gland boundaries. And then finally, we will fill the regions enclosed by the boundary to get the segmented glands. The rest of the paper is organized as follows: We describe the proposed method in detail in section II and the experimental results in section III and finally conclude the report in section IV.

[1] IISc Bangalore   [2] NIT Rourkela   [3] Jadavpur University

### A. Related Works

There has been significant research in the field of automatic gland segmentation in the past few years. In this section we will describe briefly some of the existing methods along with their merits and demerits.

Wu et al. [2] presented a region growing based approach to segment the glands from H&E stained images. They first threshold the image and then use the central lumen area to get the seed points for their region growing algorithm. This method works well for regular-shaped glands but fails for deformed malignant glands.

Farjam et al. [3] use textural features to cluster the lumen, stroma and nucleus. Then they separate out the regions containing the nucleus and remove those containing lumen and stroma.

Gunduz-Demir et al. [4] approached the problem in a slightly different way. They represented each tissue component as an object. These objects were then represented as the vertices of a graph. Then using graph connectivity, they identified the glandular regions.

All these methods work well for regular-shaped gland but fail for deformed glands. Moreover, those implementing region growing based approaches need prior information about the seed points and those methods incur heavy computational cost.

## II. METHOD

Fig 2 shows the block diagram of the proposed segmentation method. It comprises of mainly three phases: The first phase will obtain the epithelial cells from the original input image. The next phase is to classify the epithelial cells into the ones present in stromal region or in the gland boundary. Once we get the boundary epithelial cells, the third phase constructs the boundary of each gland. There are two types of glands, one having thin uniform boundary and the other having thick outer boundary. The method for boundary construction is different for both these type of glands. We get the segmented glands after post processing the boundary constructed images. The phases are described in detail in the subsections below.

### A. Segmentation of Epithelial Cells

To get the gland boundary, we need to segment the epithelial cells. The epithelial cells are darker than the rest of the components present in the tissue image. For segmentation, we perform multiple thresholding using Otsu's method[1]. The grey image is divided into five segments and the segment which has the lowest value of pixel intensities is selected as the one belonging to the epithelial cells. Let this image be 'T'.

### B. Classification of Epithelial Cells

Here, we have used the Random Forest classifier for classifying stromal and border epithelial nuclei.

**1) Random Forest**: Random forests methodology (RF)

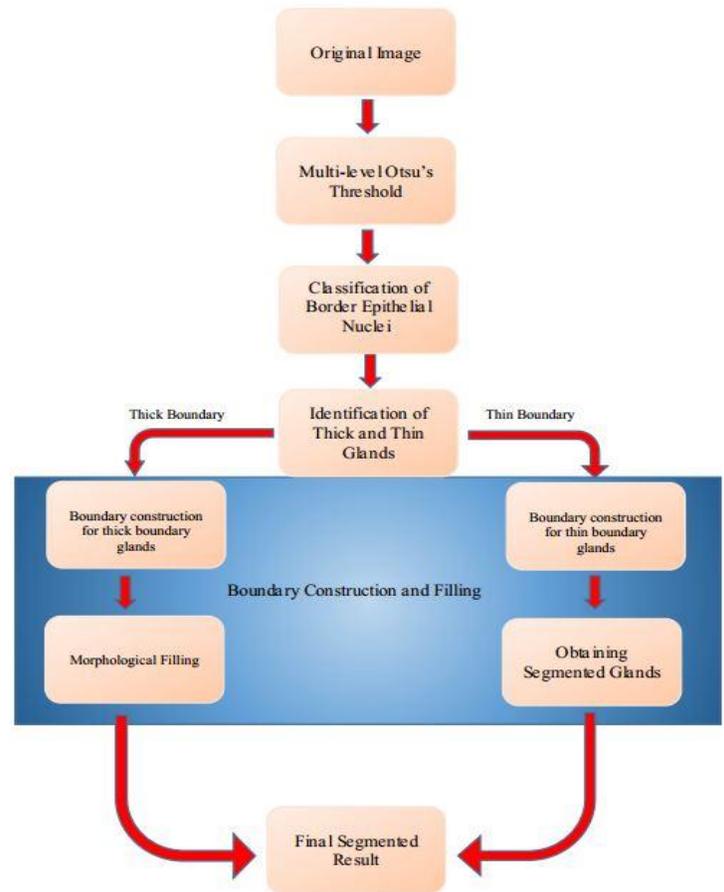

Fig 2: Block Diagram of proposed method.

was proposed by Breiman[5]. The methodology depends mainly on an ensemble of classification methods. They are an ensemble of decision trees, each one with a different subset of the training data, obtained through bootstrapping. The main advantage of this machine learning technique is that its trees are all randomly different, providing high non correlation among them, generalization and robustness. Each tree contains a collection of nodes and edges, similar to a graph.

Let the forest be composed of N number of random trees. We take S number of feature vectors each with dimension d for training the forest. Now, for each such random tree, s number of training feature vectors are sampled from S (s ⊆ S) with replacement. Let, the kth feature vector be denoted by F(k). Then, at each node of the tree, f random features chosen from F(k). A feature is chosen out of these f features based on some objective function to provide the best split on the node and consequently, children are created. There are various functions to measure the quality of a split, such as the information gain (entropy measure) or the Gini impurity. Here, we take a relative information measure for choosing the split point. This process is repeated for each subsequent node until the growth process of the tree is terminated. After the completion of the training, when a test data is entered into the classifier system, it is run down all of the trees and for each tree the data reaches a leaf node. At each such leaf node, the class probability of the input test data is evaluated. The classification is performed based on the average or weighted class probabilities of the input test data.

$$p(c|s) = \frac{1}{N}\sum_{t=1}^{N} p_t(c|s) \qquad (1)$$

In (1) N represents the total number of trees used and $p_t$ the conditional probability of class c given a vector s in each tree. Finally, the class estimate is the most probable class, i.e. $\hat{c} = \arg max_c \ p(c|x)$

**2) Feature Extraction:** We perform a connected component labelling on the segmented image(T) and then take a z by z window (z >=16) across the centroid of each of the connected components. Histogram and Haralick texture features[6] are extracted from red, green and blue channels of the original image which falls in the selected window. The window may be a part of the gland or a part of the stroma depending on whether the epithelial cell is a part of the border of the gland or if it is present in the stroma. If the centroid of the cell is marked as part of a gland in the ground truth image, then the corresponding label is 1 otherwise it is 0. These features from the training images and the train labels are used to train a random forest classifier. The histogram comprises of 32 bins for each of the channels and 13 Haralick texture features are obtained for each of the channels. Thus for each window, we get a 135 dimensional feature vector. The number of trees we used for training was N with f features for each node.

**3) Classification:** For each image in the test dataset (A and B), the feature was extracted using the same method described above and was tested using the trained Random Forest classifier. This gave us the image having the boundary epithelial nucleus. Let us call this image 'C'.

### C. Construction of Gland Boundaries

Prior to construction of gland boundaries, the images were classified into thick and thin boundary glands. This was done as follows:

**1) Classification of Thick vs Thin Boundary Gland**: There are mainly two categories of gland boundaries present. One consisting of uniform and thin epithelial cells and other set has more agglomerated epithelial cells on the gland boundary.

In order to differentiate between them, the following method is employed:

Select the binary image T and perform morphological thinning operation on it. This produces several thin edges in the image. Calculate the end points of these edges and for each such end points find all the neighbouring end points corresponding to another edge within a radius of p pixels (p= 10). Calculate the ratio r of total number of neighbours for each end points to the total number of end points in the image. A threshold ratio $N_{th}$ is calculated by applying the same method on the training images.

$$N_{th} = \frac{1}{N_{img}}\sum_{i=1}^{N_{img}} \left( \frac{1}{N_{ep}^{i}}\sum_{j=1}^{N_{ep}^{i}} \lambda_j^i \right) \qquad (2)$$

Where $N_{img}$ is the total number of training images, $N_{ep}^{i}$ is the number of end points in the $i^{th}$ image and $\lambda_j^i$ gives the number of neighboring end points for the $j^{th}$ end point in the $i^{th}$ image. If r < threshold $N_{th}$, then the image consists of thin gland boundaries else dense gland boundary is present.

**2) Constructing Gland Boundaries for Thick Images:** In this case, we propose a novel approach to connect together the bordering epithelial cells in glands having a thick boundary. To achieve this, we perform a number of line connections as per the procedure mentioned below-

First of all, Sobel operator was applied to the binary image which contains those epithelial cells which have been classified as belonging to a gland border. We get the gradient magnitude and direction at each pixel. The gradient direction is normal to the surface of the segmented epithelial cells. Find those pixels which have non zero gradient magnitude. These pixels lie on the edge of the segmented epithelial cells and the gradient direction at these pixels is normal to the surface of the epithelial cells in the outward direction. Then we divided the 2-D plane into 8 bins oriented in different directions.For each pixel with non-zero gradient magnitude, we found the bin into which the gradient direction falls. Then we take a W by W window at the starting pixel and find the mean of the pixel values of the edge preserved smoothened image based on Perona-Malik anisotropic diffusion model[7]of the red channel corresponding to the window. Let us call this mean m1. We then proceed in the direction about which the bin is centered and take another W by W window over this pixel and find the mean of the pixel values of the anisotropically diffused image corresponding to this window. Let this mean value be m2.

$$if \quad |m1 - m2| < k, \quad k \text{ is a preset threshold value.}$$

then proceed to the next pixel depending upon the direction about which the bin is centered. Else, we break the further movement.

After the line connections are made, the bordering epithelial cells have been connected together to form closed gland segments. There may however also be some stray lines present in this new connected image. To remove these stray lines a 'majority' morphological operation is done which removes most of the stray lines.

After getting the closed gland regions, we only need to 'fill' these regions. This filling operation completes the major part of the segmentation process for glands which have dense and thick boundaries.In the post processing step for thick gland boundary glands, we performed area filtration, in which small area components were removed.

**3) Constructing Gland Boundaries for Thin Images:** The proposed method to segment glands with thin boundary is similar to the one used to identify thin and thick boundaries. Here we utilize both the binary images before classification step, T and after classification, C.

Perform morphological thinning on both the images and get the corresponding end points of edges obtained from thinned image. For each end point obtained from the binary image C, search neighborhood endpoints of the image T up to a radius of p2 pixels (p2 = 20). Draw lines between the neighboring end points obtained on a binary image. Append the new line drawn image to the thresholded image C. Perform the process repeatedly up to n times (n = 5). After completion, this will connect all the boundary epithelial cells and the stromal epithelial cells in the image by forming a thick mesh in the outside portion of the gland. Holes are produced where the gland is present. Holes are also present in resulting image where neither stromal or boundary epithelial cells are seen.

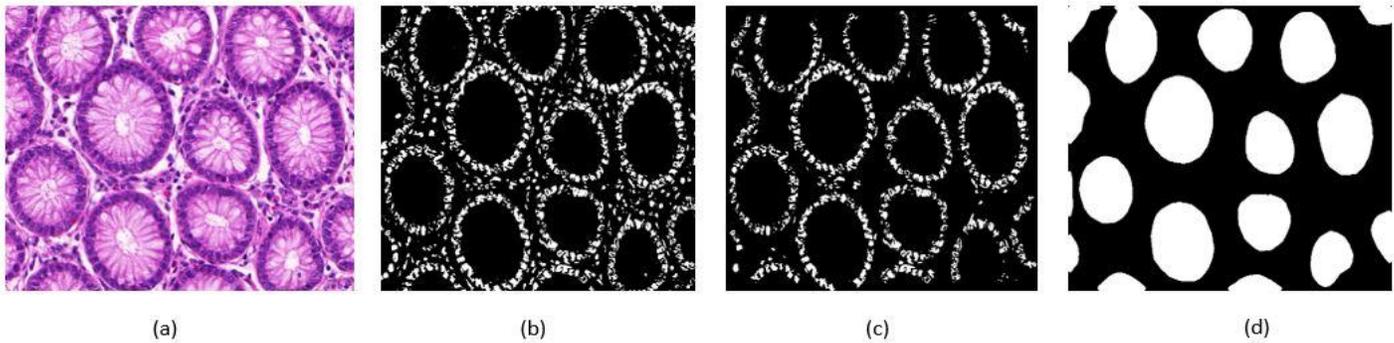

Fig 3: Shows the steps of segmentation process for thin boundary glands. (a) Original Image, (b) Nuclei segmented image, (c) Classified Image containing border epithelial cells, (d) Final Segmentation Result

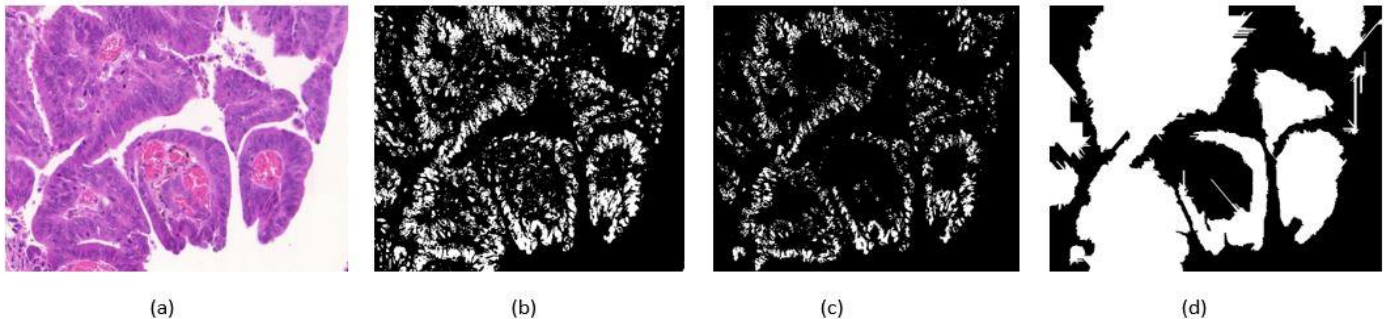

Fig 4: Shows the steps of segmentation process for thick boundary glands. (a) Original Image, (b) Nuclei segmented image, (c) Classified Image containing border epithelial cells, (d) Final Segmentation Result

These holes are filled and the corresponding image containing the gland is obtained after post processing steps.

The post processing steps include identifying gland regions from the set of gland and non-gland regions. This is obtained by detecting whether the filled regions are surrounded by boundary epithelial cells or stromal cells with the help of classified binary image C.

## III. EXPERIMENTS AND RESULTS

### A. Dataset

The dataset we used for our experiment is Warwick-QU available at the Warwick University, GlaS challenge website[8]. The dataset consists of 165 H&E images derived from 16 H&E stained histological sections of stage T3 or T4 colorectal adenocarcinoma. It was divided into one training set and two test set: TestA and TestB. Table 1 depicts the distribution of training and test data used for experiments. Out of the set of 165 images 85 images were used for training and 80 images were used for testing. Each section belongs to a different patient, and sections were processed in the laboratory on different occasions. Thus, the dataset exhibits high inter-subject variability in both stain distribution and tissue architecture.

Table 1: Distribution of images in training and test data

| DATASET | NO. OF IMAGES |
|---|---|
| TRAIN | 85 |
| TEST A | 60 |
| TEST B | 20 |

The digitization of these histological sections into whole slide images (WSIs) was accomplished using a Zeiss MIRAX MIDI Slide Scanner with a pixel resolution of 0.465µm. The WSIs were subsequently rescaled to a pixel resolution of 0.620µm (equivalent to 20x objective magnification).

### B. Implementation Parameters

A window of size with $z=24$ was chosen around the epithelial cell because it was sufficient to cover the entire cell and provide sufficient textural and intensity features. A random forest of 500 trees was trained with 20 features randomly chosen for each node. For boundary construction in thick glands, a window size of $w=5$ was chosen so as to average out the effect of any spurious color changes. The decision threshold 'k' used to decide the further propagation of line was chosen to be 45 based on domain knowledge. Table 2 lists the various parameters and their associated values which we have used in the segmentation process.

Table 2: Values of various parameters used in the segmentation process

| Parameter | Value |
|---|---|
| z | 24 |
| W | 5 |
| N | 500 |
| f | 20 |
| k | 45 |

### C. Performance Measure and Comparisons

Features were extracted from 85 training images and a random forest classifier was trained using these features. During the testing process features were again extracted from the test images and the already trained random forest classifier was used to classify the epithelial cells present in the images as either belonging to gland boundary or as present in the stroma. After classification, we proceeded with our gland segmentation process. In this section we compare our results with some of the existing methods. Fig 3(a) and 4(a) shows the original image. Fig 3(b) and 4(b) shows the segmented binary image containing both stromal and boundary epithelial nuclei. Fig 3(c) and 4(c) shows the segmented image containing boundary epithelial nuclei after classification using Random Forest. Fig 3(d) and 4(d) shows the final segmented image after boundary construction.

For our comparisons we use the existing metrics like F1-Score, Object Dice and Object Hausdorff.

**F1Score**- This measures whether the detected gland is truly a gland or not.

$$F1Score = \frac{2 * Precision * Recall}{Precision + Recall}$$

Where, $Precision = \frac{TP}{TP+FP}$ , $Recall = \frac{TP}{TP+FN}$ , TP=True Positive, FP=False Positive, FN=False Negative.

**Object Dice**- This measures the extent of intersection in the segmented gland with the ground truth object.

**Object Hausdorff**- This measure the similarity in shape between the segmented gland and ground truth object.

The scores are calculated for images in the test set A and B provided in the Warwick dataset. Table 3 shows scores obtained by our method and also that obtained by various other teams using the same dataset. Our results show consistency irrespective of the shape and size of the glands. The proposed method can be implemented using few training examples due to the employment of Random Forest classifier. Fig 5 shows the comparison of segmentation result with the ground truth for some sample images. Column (a) represents the original image. Column (b) represents the ground truth and column (c) shows the segmented results obtained from our proposed method. The colored regions in the images shown in columns (b) and (c) are the gland regions. The images in the first two rows have been processed by considering them to have thin boundary glands. The image in third row corresponds to one with a thick gland boundary. Fig 6 shows the results for few of the other sample images. 1st row has the original images. 2nd row has the ground truth for segmentation and 3rd row shows the segmentation results obtained using the proposed method.

### D. Discussion

Our results are consistent for both benign as well as malignant glands. The proposed method can be implemented using few training examples due to the employment of Random Forest classifier. The proposed method suffers from some problems which affect its performance. If two thick glands are very close together and there is no contrasting boundary between them, then the two glands may be connected together and it will be segmented as one. In case of thin glands, the outer boundary is absent in the final segmented result.

Table 3: This table compares the results of our proposed method with the results of some of the best methods presented in the GlaS Challenge, 2015.

| TEAMS | F1-SCORE A | B | OBJECT DICE A | B | OBJECT HAUSDORFF A | B |
|---|---|---|---|---|---|---|
| BIOIMAGE INFORMATICS TEAM (GLAS CHALLENGE) | 0.45 | 0.35 | 0.5 | 0.65 | 275 | 210 |
| SUTECH (GLAS CHALLENGE) | 0.5 | 0.15 | 0.6 | 0.5 | 180 | 260 |
| CVIP DUNDEE (GLAS CHALLENGE) | 0.8 | 0.62 | 0.85 | 0.7 | 60 | 210 |
| VISION4GLAS (GLAS CHALLENGE) | 0.65 | 0.51 | 0.72 | 0.62 | 105 | 210 |
| PROPOSED METHOD | 0.54 | 0.52 | 0.65 | 0.57 | 126 | 262 |

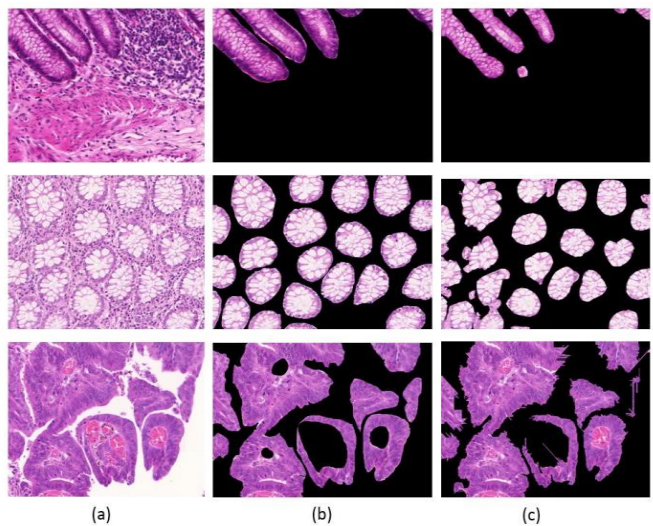

(a)  (b)  (c)

Fig 5: The comparison of segmentation results for some sample images with the ground truth (a) Original image (b) Ground truth (c) segmentation result by proposed method

### IV. Conclusion

In this paper, we provide a method to separate the epithelial nucleus from the stromal nucleus using Random Forest classifier and then implement a novel boundary construction approach to obtain the glands separately. The use of Random Forest makes the classification of boundary epithelial cells efficient even if we have lesser number of data available for training. In future, we want to improve our algorithm so as to provide better accuracy for closely spaced glands with thick boundary. Moreover, there is also scope for further improvement in the feature selection for classification of boundary epithelial cells.

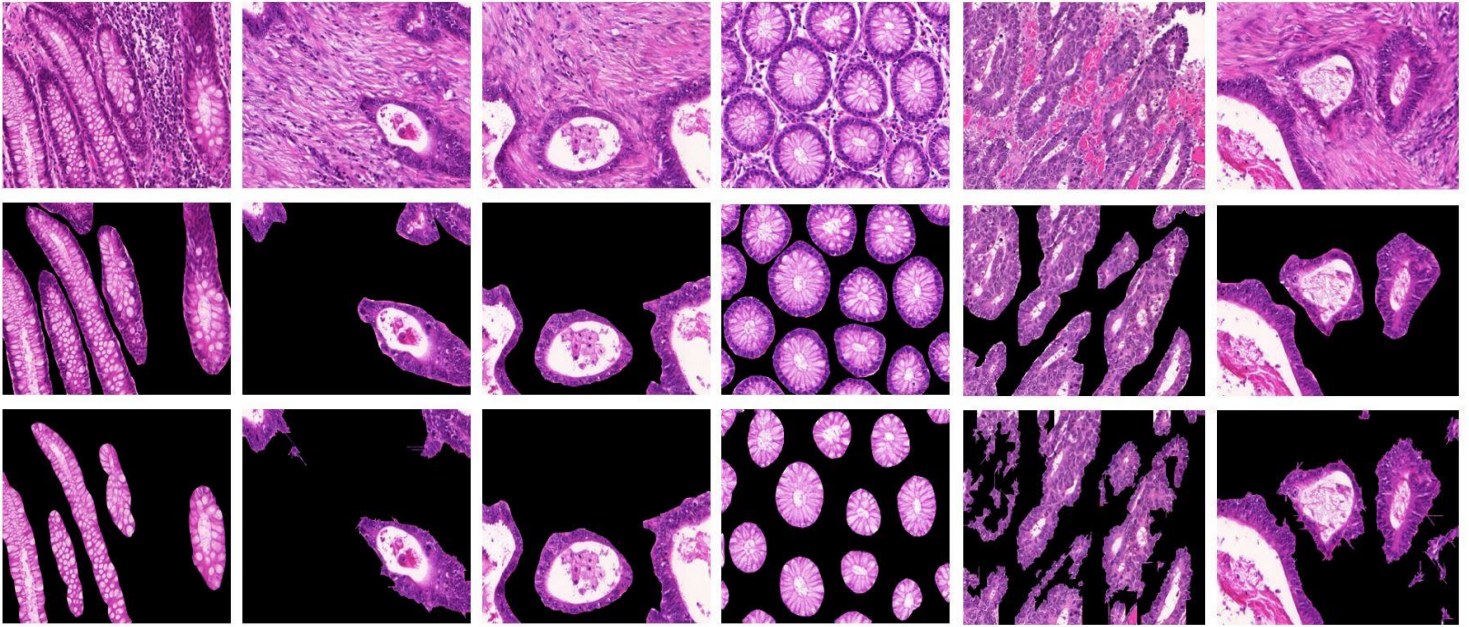

Fig 8: Comparison of segmentation results for some sample images with the ground truth. 1ˢᵗ row- Original image 2ⁿᵈ row – ground truth 3ʳᵈ row – Segmentation results obtained from proposed method.